\documentclass[letterpaper, 10 pt, conference]{ieeeconf}  

\IEEEoverridecommandlockouts                              

\usepackage{amsmath,amssymb,amsfonts}
\usepackage{graphicx}
\usepackage{textcomp}

\usepackage{overpic}
\usepackage[font=footnotesize]{subcaption}
\usepackage{tikz}
\usepackage{units}
\usepackage{txfonts}
\let\mathbb=\varmathbb

\usepackage{booktabs}
\usepackage{tabularx}
\usepackage{textcomp}
\PassOptionsToPackage{hyphens}{url}
\usepackage[hidelinks]{hyperref}

\usepackage{wallpaper}
\definecolor{colorPRDarkBlue}{HTML}{0F4176}
\ULCornerWallPaper{1.0}{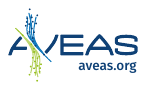}
\usepackage{fancyhdr}
\newcommand\renderTitleDOI[2]{#1, doi: \href{https://doi.org/#2}{#2}}

\title{\LARGE \bf
An Approach to Systematic Data Acquisition and Data-Driven Simulation for the Safety Testing of Automated Driving Functions$^*$
}

\author{%
L.\,Eisemann$^{1,7}$, 
M.\,Fehling-Kaschek$^{2\text{A}}$, 
H.\,Gommel$^{3}$,
D.\,Hermann$^{4,7}$,
M.\,Klemp$^{5}$,
M.\,Lauer$^{5}$,
B.\,Lickert$^{2\text{A}}$,\\
F.\,Luettner$^{2\text{A}}$, 
R.\,Moss$^{2\text{A}}$, 
N.\,Neis$^{5,7}$, 
M.\,Pohle$^{2\text{B}}$,
S.\,Romanski$^{6}$,
D.\,Stadler$^{2\text{C},5}$, 
A.\,Stolz$^{2\text{A}}$, 
J.\,Ziehn$^{2\text{C}}$,
J.\,Zhou$^{5,7}$%
\thanks{*This publication was written in the context of the AVEAS research project (www.aveas.org), funded by the German Federal Ministry for Economic Affairs and Climate Action (BMWK) within the program ``New Vehicle and System Technologies''.}
\thanks{\raggedright$^{1}$Stuttgart Media University, 70569, Stuttgart, Germany} 
\thanks{\raggedright$^{2}$Fraunhofer-Gesellschaft e.V., 80007, Munich, Germany}%
\thanks{~~~~$^{\text{A}}$Fraunhofer EMI, 79588 Efringen-Kirchen, Germany}%
\thanks{~~~~$^{\text{B}}$Fraunhofer IVI, 01069 Dresden, Germany}%
\thanks{~~~~$^{\text{C}}$Fraunhofer IOSB, 76131 Karlsruhe, Germany}%
\thanks{\texttt{\{mirjam.fehling-kaschek@emi, florian.luettner@emi, maria.pohle@ivi, jens.ziehn@iosb\}{}.fraunhofer.de}}%
\thanks{\raggedright$^{3}$GOTECH GmbH, 71287 Weissach, Germany}%
\thanks{\raggedright$^{4}$Technische Universität München, 85748 Garching b. München \texttt{david.hermann@tum.de}}%
\thanks{\raggedright$^{5}$Karlsruhe Institute of Technology, 76131, Karlsruhe, Germany}%
\thanks{\raggedright$^{6}$understandAI GmbH, 76227, Karlsruhe, Germany}%
\thanks{\raggedright$^{7}$Porsche Engineering Group GmbH, 71287, Weissach, Germany}
}

\newcommand\subs[1]{_{\text{#1}}}

\begin{document}

\maketitle
\thispagestyle{fancy}

\begin{abstract}
With growing complexity and criticality of automated driving functions in road traffic and their operational design domains (ODD), there is increasing demand for covering significant proportions of development, validation, and verification in virtual environments and through simulation models.

If, however, simulations are meant not only to augment real-world experiments, but to replace them, quantitative approaches are required that measure to what degree and under which preconditions simulation models adequately represent reality, and thus, using their results accordingly. Especially in R\&D areas related to the safety impact of the ``open world'', there is a significant shortage of real-world data to parameterize and/or validate simulations---especially with respect to the behavior of human traffic participants, whom automated driving functions will meet in mixed traffic.

We present an approach to systematically acquire data in public traffic by heterogeneous means, transform it into a unified representation, and use it to automatically parameterize traffic behavior models for use in data-driven virtual validation of automated driving functions.
\end{abstract}

\section{INTRODUCTION}\label{sec:intro}

\PARstart{T}{he} introduction of automated driving functions achieving SAE Level 3 and above presents a significant shift in the requirements for validation and testing approaches, due to a substantial increase in complexity. This complexity is a combination of several related factors: The increased scope of the operational design domain (ODD) that needs to be covered, including diverse urban scenarios; the increased responsibility of the automated driving functions for a wider range of aspects within the dynamic driving task (DDT); the increased level of safety that must be achieved without relying on the human driver as a fallback; and the increased complexity of the hardware/software system capable of operating under these conditions. The latter typically involves a wide array of heterogeneous sensors, different high-performance hardware platforms interoperating in realtime, and a combination of classically-developed software and modern machine learning (ML) models for various tasks that require new development and testing paradigms.

The increase in complexity is mirrored in the evolutions of automotive safety standards from functional safety (ISO~26262) over safety of the intended functionality (ISO~21448, SOTIF) to safety and artificial intelligence (ISO~8800). This evolution marks a transition from the consideration of individual components under general conditions towards the consideration of the overall system (including, e.g., training data for ML) in its realistic environment (including, e.g., the behavior of human traffic participants \cite{schlag2016automatisiertes} and passengers of automated vehicles \cite{konig2015nutzergerechte}). At the same time it highlights the necessity of addressing not only ``known'' limitations and failure modes, but also defining processes to identify system limits that were not known or specified during development time. The combination of these effects manifested notably in the fatal 2016 accident between a Tesla Model S and a  tractor-semitrailer truck in Florida \cite{ntsb2016tesla}.

\subsection{The Role of Data in Validation}

For validation approaches of automated driving functions that aim at establishing quantitative estimates for the safety impact of operation, the availability of statistically significant real world data is crucial: Any tested scenario, especially a scenario where the \emph{system under test} (SUT) fails, must be quantified in a stochastic sense to determine the impact of this failure on the safety level. Failure scenarios that have a high exposure (e.g., failing to detect a leading vehicle) will be considered a priority in development, while scenarios that are (based on available data) extremely unlikely (e.g., failing to detect a zebra on a zebra crossing) may be considered low priority or even be left as residual risks. Specifically, the perspective would be that if the SUT ever has an accident, it will be considered ``well designed'' if the accident occurs within the stochastic rate predicted during development and under conditions that were known during development. Achieving this degree of accuracy by means of virtual validation requires the provision of real-world data to parameterize state of the art behavior models (cf. also \cite{neis-literaturereview}).

\renewcommand\tabularxcolumn[1]{m{#1}}
\newcolumntype{Y}{>{\footnotesize\raggedright\arraybackslash}X}
\newcolumntype{Z}{>{\footnotesize\centering\arraybackslash}X}
\newcolumntype{L}[1]{>{\footnotesize\raggedright\let\newline\\\arraybackslash\hspace{0pt}}m{#1}}

\begin{table*}
\caption{Overview of different employed acquisition methods and distinctive properties to be considered in the unified format.}
\label{tab:acquisition-methods}
\begin{tabularx}{\textwidth}{@{}YYZZZZYZL{2.4cm}@{}}
\toprule
Platform / Source & Environment Sensor(s) & Ground FOV & Interior Observation & Platform Speed & Max. Rec. Duration & Mapping Principle & Notable \emph{missing} features & Notable Acq. Limitations \\ \midrule
Road vehicle~\cite{Haselberger-Jupiter-2022} & LiDAR, RGB camera & $\approx 50~\text{m}^2$ & --- & 0--200 km/h and above & 5 h & LiDAR-based & --- & Strongly adverse driving conditions \\
Infrastructure & LiDAR & $\approx 100~\text{m}^2$ & --- & --- & weeks & landmark-based ref. & lights, colors & Strong precipitation \\
Infrastructure & LWIR camera & $\approx 100~\text{m}^2$ & --- & --- & weeks & landmark-based ref. & lights, colors & Temp. below freezing \\
Aircraft & RGB camera & $\approx 400~\text{m}^2$ & --- & 70--200\,km/h & 2 h & orthophoto-based & lights, small VRUs & Strong precipitation, night, fog \\
GIDAS data & (accident reconst.) & --- & (accident reconst.) & --- & 30 s prior to accident & manual & accurate trajectories & Only selected accidents \\
Driving sim. studies & sim. ground truth & --- & mult. driving/human param. & --- & 1 h & Sim. ground truth & --- & Only one instructed test person per run \\ \bottomrule
\end{tabularx}
\end{table*}

\subsection{Scope of the Paper}

In this paper, we present a pipeline from data acquisition to the optimized parameterization of behavior models and realistic traffic simulations of relevant use cases. First, we describe the steps which need to be taken to process data recordings to obtain trajectories with relevant parameters of all recorded traffic participants, accompanied by OpenDRIVE-based road representations \cite{asam-opendrive-2022}. Second, we introduce an approach to derive optimized behavior models for PTV Vissim \cite{Vissim2023} based on the trajectories obtained from the recorded data. We present an exemplary optimization of the desired speed distributions of cars and trucks based on data collected via aerial image acquisition as described before. We also discuss how the optimization can be extended in the future. Finally, we describe the usage of the derived optimized behavior models for traffic simulations of relevant use cases. Here, PTV Vissim is coupled with a Carla-based \cite{pmlr-v78-dosovitskiy17a} simulator to generate a digital twin of the SUT \cite{Haselberger-Jupiter-2022}.

\section{MEANS OF DATA ACQUISITION}\label{sec:data-aquisition}

Within the scope of our studies, various methods of traffic data acquisition are employed, each with individual strengths and weaknesses, namely road vehicle-based, infrastructure-based, and aircraft-based platforms featuring either RGB cameras, LWIR cameras, LiDAR, radar, or a combination thereof (cf. Tab.~\ref{tab:acquisition-methods}). Compatible data are also acquired from driving simulator studies and accident reconstructions, leading to a wide range of different data granularities and available fields (see Tab.~\ref{tab:acquisition-methods} for an overview). Each method includes a highly individual processing chain based on available sensor data and platform parameters, whose output, however, is a unified format of \emph{map-referenced trajectories}. This format comprises a minimal set of mandatory parameters that can be determined across any acquisition method, sufficient to formulate, parameterize and/or evaluate driving behavior models. These parameters include:
\begin{itemize}
\item a map representation based on OpenDRIVE \cite{asam-opendrive-2022}, including road geometry, relevant traffic signs, etc.
\item dynamic traffic participants, containing at least:
\begin{itemize}
    \item a coarse classification of the participant (car, truck bicycle, pedestrian, ...);
    \item its physical size (at least length and width);
    \item its trajectory (geographic \emph{and} road map coordinates over time) at a rate of several Hz (pref.: $>$ 10 Hz);
    \item uncertainty specifications for all measured and derived values, limited to Gaussians for real-valued data and probability mass functions for finite, discrete properties (classes, binary states, etc.);
\end{itemize}
\item unobserved areas due to occlusions or sensor limits;
\item metadata such as timestamp, geographic location, etc., and information whether the scenario is entirely ``natural'', or partly or completely ``staged'' or ``simulated'';
\item derived information from post-processing such as time to collision (TTC) \cite{westhofen2023criticality} and post encroachment time (PET) \cite{peesapati2018can_PET}; accelerations per participant over time; traffic density.
\end{itemize}
A wider range of optional parameters cannot be extracted by all methods, but may be valuable for specific models or hypotheses, and is in some cases added by fusion with external sources, including weather and lighting conditions, vehicle light states, road conditions, high-resolution lateral positions (cf. \cite{neis-lateral}) and (ego) vehicle interior properties such as driver pose and attention direction, which is used, e.g., in the SCM behavior simulation model \cite{fries2022driver}. Within the driving simulation studies, multiple objective (technical/human) and subjective measurements will be derived using a virtual reality-based advanced driving simulation system as depicted in \cite{bouchner2016interactive} using different interior setups, one of them based on the vehicle used for data acquisition.

\subsection{Scope and Assumptions for the Acquisition}\label{sec:data-acquisition_scope}

Regions of interest for data acquisition range from highways over urban environments to parking areas. The scope of the project is the acquisition of \emph{critical} scenarios for the validation of near-future automated driving functions in mixed traffic, based on the perspective of ISO~21448, and considers relevant factors to comprise...
\begin{itemize}
    \item criticality introduced by the behavior of human traffic participants as statistically observable today;
    \item criticality introduced by characteristics of automated vehicles' sense--plan--act pipelines through which their behavior deviates from  human behavior while in automated driving mode (such as sensor characteristics, ML perception and fusion \cite{zhou-unknown}, planning methods, ...);
    \item criticality from the transition between manual and automated control of the DDT, either by handover from the system to the human driver, or from intervention of the human driver into the system's actions.
\end{itemize}
Therefore, accident databases and police records were filtered by relevant accident types for automated driving functions, to determine specific acquisition sites with a relatively high exposure of homogeneous such accident types. At these sites, the goal is to acquire not the relatively rare actual accidents, but a larger proportion of near misses. Here, we define \emph{near misses} as ``scenarios in which an accident was averted predominantly due to stochastic effects, not (exclusively) due to the controlled behavior of the involved parties''. This definition enables the stochastic verification for derived models: A variation of a ``near miss'' scenario under stochastic parameters determined from the data-driven approach should approximate the actual accident statistics.


\subsection{Map-Referencing}\label{sec:map-referencing}

To enable meaningful behavior analyses on the detected traffic participants, their trajectories must be viewed in relation to the underlying road network geometry and topology. Whether, for example, two cars are on the same lane or on neighboring lanes is decisive in established behavior models like the one used in PTV Vissim \cite{Vissim2023}, but can only be determined reliably by considering the road geometry. Similarly, behaviors such as lane changes depend on available options (available lanes) and demands (such as upcoming highway exits or speed limits).

Hence, the observed traffic participants and their trajectories must be referenced with detailed map data, and in particular, this map data must above all assure that the dynamic objects (traffic participants) and the map data are consistent with each other. The geographic accuracy of the maps is, in contrast, an expendable feature. This consideration is of highly practical importance: Mapping and object detection may be performed by different methods, on different data and---possibly---at different times using different platforms. Furthermore, the ASAM OpenDRIVE format (as the most widely adopted map standard in virtual testing, see \cite{chiang_automated_2022}) uses Cartesian coordinates generally disregarding earth's curvature. Hence, in addition to measurement errors, surface projection errors can be significant.
This is not a relevant limitation either for pure simulation scenarios or automated driving where the vehicle performs active ego localization within the map. When merging data from different sources, however, offsets from mapping uncertainty, reprojection uncertainty and localization/detection uncertainty (generally independent) will easily accumulate to orders of magnitude comparable to typical lane widths. 

Therefore, to primarily assure that extracted map data and extracted traffic data are consistent, our approach requires each acquisition method to provide elementary mapping directly related to the raw sensor data (opposed to requiring purely a detection of dynamic objects which are overlaid with map data from a different source), as described in \cite{eisemann-opendrive}.

\begin{figure}
    \centering
    \begin{subfigure}{\columnwidth}
    \includegraphics[width=\columnwidth]{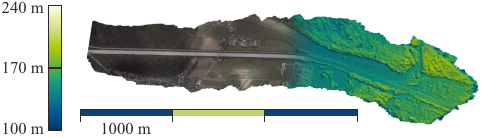}
    \caption{Computed orthophoto and elevation map}
    \vspace{2mm}
    \label{fig:aerial-ortho}
    \end{subfigure}\\
    \begin{subfigure}{\columnwidth}
    \includegraphics[width=\columnwidth]{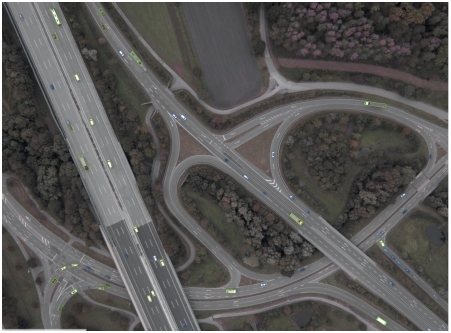}
    \vspace{-5mm}
    \caption{Detection of driving and parked cars (blue) and trucks (green)}
    \vspace{2mm}
    \label{fig:aerial-detection}
    \end{subfigure}\\
     \begin{subfigure}{\columnwidth}
    \includegraphics[width=\columnwidth]{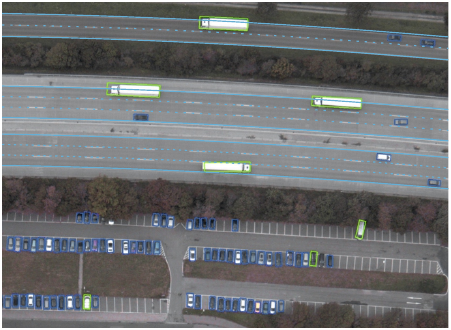}
    \vspace{-5mm}
    \caption{Detected and tracked vehicles with OpenDRIVE lanes (light blue)}
    \label{fig:aerial-statdyn}
    \end{subfigure}\\
    \caption{Samples of the aerial image traffic data acquisition method, based on images recorded by a Flight Design CTSW ultralight aircraft between the cities of Karlsruhe and Rastatt.}
    \label{fig:aerial}
\end{figure}

\subsection{Object Detection and Tracking}

\begin{figure}
    \centering
    \includegraphics[width=0.49\textwidth]{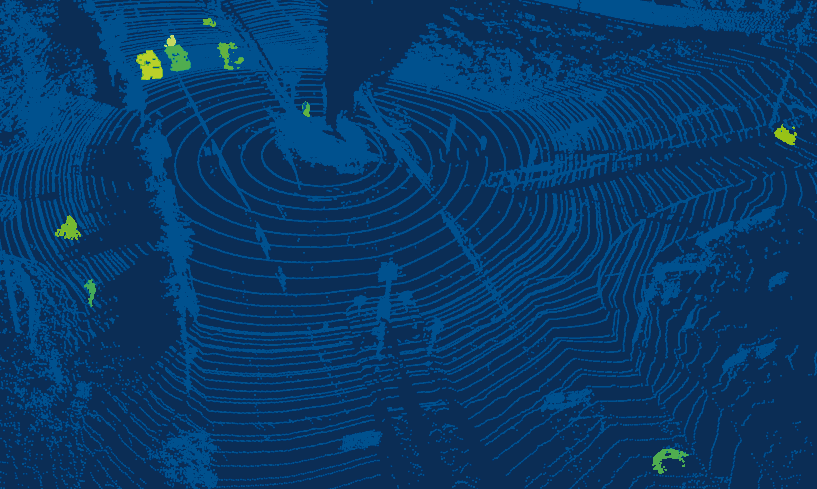}
    \caption{A point cloud from the infrastructure recording after applying the processing pipeline. Background points are colored in blue and points belonging to an object have a different color based on the tracked object.}
    \label{fig:kit_lidar}
\end{figure}

We describe the process of object detection and tracking on the example of the aerial image acquisition method shown in Fig.~\ref{fig:aerial} and used subsequently in Sec.~\ref{sec:model-parametrization}, as well as for the LiDAR / infrastructure-based acquisition shown in Fig.~\ref{fig:kit_lidar} and briefly for the road vehicle-based acquisition.

\subsubsection{Aerial Data Acquisition}

The data are acquired by a Flight Design CTSW ultralight aircraft, carrying a sensor unit combining a calibrated 4K camera with IMU and GNSS. Images are acquired at 20 FPS and a ground sampling distance of approximately 12.5~cm.

In the first processing step, a digital elevation model (DEM) is computed through bundle adjustment, using raw images and the IMU/GNSS data (Fig.~\ref{fig:aerial-ortho}). This step assures a robust registration of the raw images into an \emph{orthophoto}, a projection of the acquired data onto reference ground plane coordinates (in particular eliminating geometric displacement effects from elevated structures), and hence an intrinsically consistent geo-referencing within the raw data footage as outlined in Sec.~\ref{sec:map-referencing}. The orthophoto enables an easy extraction of the road geometry (in the OpenDRIVE format). For the shown dataset, this information was annotated manually; for large-scale acquisition of different regions, an automatic implementation is currently being developed.

Furthermore, object detection is performed on the raw data stream using the CNN-based rotation-invariant detection architecture ReDet \cite{han2021redet}, available in MMRotate \cite{mmrotate}, trained on the iSAID \cite{waqas2019isaid} dataset and fine-tuned on 1324 manually annotated images from the target domain. The iSAID instance segmentation dataset provides a wide range of classes over 2806 high-resolution images, of which only the distinction between cars (iSAID class ``small vehicle'') and trucks (iSAID class ``large vehicle'') is relevant for the use case. The detection already yields object-aligned bounding boxes for each vehicle, providing an indication of the expected direction of travel (Fig.~\ref{fig:aerial-detection}). Detection results per vehicle class are shown in Tab.~\ref{tab:aerial-results} using 510 
manually annotated test images (mean average precision (mAP) is obtained at an IoU threshold of 0.5), comparing pure iSAID training with fine-tuning on the target domain. Both runs use the ReDet detector with a ReResNet50 \cite{han2021redet} backbone.

Subsequently, detected objects are tracked between the geo-referenced frames to achieve map-referenced trajectories w.r.t. the extracted OpenDRIVE maps (Fig.~\ref{fig:aerial-statdyn}), using a DeepSORT-based tracker \cite{wojke2017simple}. The original CNN for appearance feature extraction is replaced with OSNet [17] which is trained on a custom built re-identification dataset based on cropped image patches from the VisDrone \cite{visdrone} and UAVDT \cite{uavdt} multi-object-tracking datasets. The custom re-identification dataset comprises a total of 31\,591 images containing 1\,372 various vehicles.

\begin{table}
\caption{Quantitative results over frames and vehicle classes of the aerial acquisition using ReDet \& ReResNet50.}
\label{tab:aerial-results}
\begin{tabularx}{\columnwidth}{@{}Xlll@{}}
\toprule
Train datasets & AP$\subs{car}$ & AP$\subs{truck}$ & mAP \\ \midrule
iSAID \cite{waqas2019isaid} & 89.6 & 69.9 & 79.7 \\
iSAID  + fine-tuning on target domain& \textbf{90.4} & \textbf{85.5} & \textbf{87.9} \\ \bottomrule
\end{tabularx}
\end{table}

\subsubsection{Infrastructure LiDAR Data Acquisition}

For the LiDAR-based infrastructure recording, the LiDAR is mounted on a pole in approx. 3~m height next to intersections selected according to Sec.~\ref{sec:data-acquisition_scope}. As the processing is done on-site, it requires efficient algorithms. In the infrastructure-based recording, we can consider the background as static, with only traffic participants moving in consecutive point clouds, enabling a background subtraction approach, based on the general idea of \cite{8818659}, in which the background is constructed based on the range of individual LiDAR beams. Accordingly, in each point cloud, we remove the background and cluster the remaining points. These detected clusters represent the traffic participants. To track them, we formulate tracking as a min-cost flow problem and build up a graph of detections over time by connecting detections in consecutive point clouds with their Euclidean distance as cost. We use the muSSP algorithm \cite{NEURIPS2019_b1d10e7b} to efficiently solve this min-cost flow problem. Fig.~\ref{fig:kit_lidar} shows the result after tracking individual traffic participants. 

\subsubsection{Vehicle-Based Data Acquisition}

The vehicle-based data acquisition follows a similar principle. The recording vehicles make use of camera and LiDAR data to use their complementary strengths. LiDARs provide accurate 3D information but lack information density in far distances in comparison to cameras. As the recording vehicle is typically moving in critical scenarios, the background is harder to detect and remove in vehicle-based data acquisition. Hence, only the ground plane is algorithmically removed from the point cloud to simplify the dynamic object detection task. The camera signal is used to detect and track dynamic objects with an ML-driven pipeline. The corresponding 3D shapes and movements are obtained by separate ML models. As the data processing is mostly performed offline, human quality reviews can elevate the data quality to ground truth accuracy where needed. Through the use of accurate GNSS/IMU systems, the 3D objects are referenced onto the individual OpenDRIVE maps.


\section{MODEL PARAMETERIZATION}\label{sec:model-parametrization}
Under the premise of evaluating automated driving functions against human traffic participants in simulations, we aim at developing an optimization framework to parameterize driving behavior models based on the previously showcased data.
These models can then be used in agent-based simulations, in which the interaction between a vehicle with automated driving functions (the SUT) and realistically behaving opponents can be observed and assessed from a safety perspective.
For data driven validation of SUTs, it is important to do so not just by replaying recorded trajectories of opponents identically, but rather reproducing their characteristics from a stochastic point of view.
As such, as part of the optimization framework, we determine the log-likelihood of recorded observations given the simulated distributions of interest, such as the vehicle velocities, as shown below.
Subsequently, the parameters of the underlying behavior models are iteratively improved such that the likelihood to observe the data, under the hypothesis that the model parameters of the simulation are true, is maximized. 
Having determined an optimized model parameterization, we can then create arbitrary variations of realistic traffic dynamics, in which automated driving functions can be tested interactively or which can be used to create alternative scenarios.

\subsection{Simulation Model Framework}
All of the simulations of traffic dynamics shown hereafter were performed using PTV Vissim \cite{Vissim2023}.
Here, the Wiedemann99 model \cite{VissimWiedemann2023} was used to simulate the vehicle following dynamics of the traffic participants. 
Each of the model's ten parameters describes a different aspect of the agent dynamics when following a preceding vehicle.
Further, a modified version of the model by Sparmann \cite{Sparmann1978} was used to characterize lane change dynamics of the vehicles.
As part of an early investigation, a simulation setup was created based on the initially recorded aerial data showcased in Sec.~\ref{sec:data-aquisition} / Fig.~\ref{fig:aerial}.
Therefore, a 3-lane 4~km long interstate highway (\emph{autobahn}) section with a traffic volume of 1680 cars and 320 trucks per hour was created within PTV Vissim.
The distribution of the \emph{desired velocity} for each vehicle type was then set based on a Gaussian distribution, with a given mean $\mu$ and standard deviation $\sigma$.
To ensure the simulated traffic dynamics have reached a stable state, an additional inflow track with the length of 1\,500~m was added to the previously defined highway (but disregarded in the later analysis of the simulation results).

\subsection{Optimization Framework}
\begin{figure*}
    \centering
    \includegraphics[width=1\textwidth]{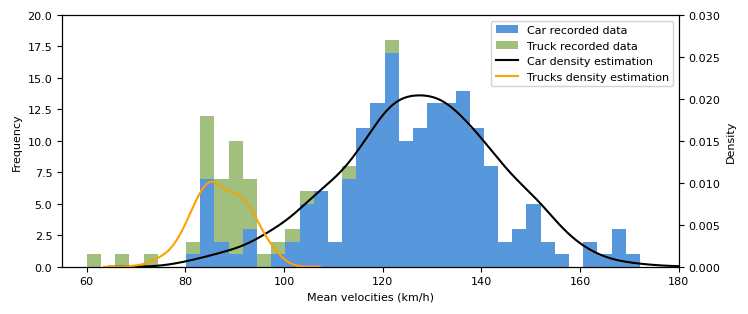}
    \caption{Comparison between the histogram of the extracted mean velocities from our recorded data (trucks: green, cars: blue) and the respective density estimations determined from the simulated mean velocities (trucks: orange, cars: black).}
    \label{fig:IV_histogram}
    \vspace{-10pt}
\end{figure*}

\begin{figure}
    \centering
    \includegraphics[width=0.49\textwidth]{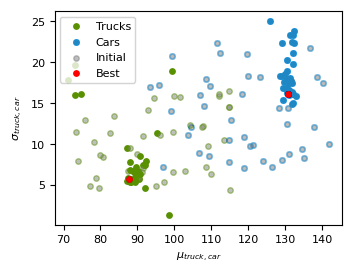}
    \caption{Comparison of the respective 100 resulting $\mu$ and $\sigma$, of each individual estimation run. The parameters for the truck type are depicted in green and car type in blue. Further, initial guesses for each estimation run are shown in gray.}
    \label{fig:IV_Parameters}
    \vspace{-10pt}
\end{figure}

To evaluate the applicability of the concept, we optimized the four parameters ($\mu\subs{vel,car}$, $\sigma\subs{vel,car}$, $\mu\subs{vel,truck}$, $\sigma\subs{vel,truck}$) of the \emph{desired velocity} distributions needed to recreate the mean recorded velocities from Sec.~\ref{sec:data-aquisition}.
This includes the data extracted from the trajectory of 178 cars and 33 trucks, recorded within a time frame of approximately 2~min.
For the optimization, as previously mentioned, we defined an objective function to be minimized via the log-likelihood ($\ln(\mathcal{L})$) based on simulated and recorded mean velocities of the respective observed vehicle types,
\begin{equation}
\begin{split}
    \!\!-\ln(\mathcal{L})= &-\!\!\sum_{k}^{N\subs{car}}\!\ln(\phi\subs{vel,car}(v\subs{car,k})) -\!\!\sum_{l}^{N\subs{truck}}\!\ln(\phi\subs{vel,truck}(v\subs{truck,l}))\,,\!
\end{split}
    \label{eq:objective}
\end{equation}
where $\phi\subs{vel,car}$ and $\phi\subs{vel, truck}$ are the simulated probability densities of car and truck velocities, $v\subs{car,k}$ and $v\subs{truck,l}$ are the recorded velocities of the $k^{\text{th}}$ car and $l^{\text{th}}$ truck, and $N\subs{car}$ and $N\subs{truck}$ denote the total number of recorded cars and trucks. The simulated probability densities $\phi\subs{vel,car}$ and $\phi\subs{vel,truck}$ 
were estimated using Gaussian kernel density estimators as implemented in the Python scikit-learn package \cite{scikit-learn}. 
The objective function was minimized using the Nelder--Mead algorithm \cite{Gao2012} implemented in the Python SciPy package \cite{2020SciPy-NMeth}.
To confirm robustness of the optimization method, the parameter estimation was repeated 100 times from different initial parameter values sampled using a uniform distribution.
The estimated \emph{desired velocity} parameters corresponding to the best objective function value of the 100 estimation runs were
 $\mu\subs{vel,car}=131.05~\text{km/h}$ and $\sigma\subs{vel,car}=17.48~\text{km/h}$ for the \emph{car} type vehicles and  $\mu\subs{vel,truck}=89.22~\textrm{km/h}$ and $\sigma\subs{vel,truck}=6.20~\text{km/h}$ for the \emph{truck} type vehicles.
A histogram of the recorded velocity data for each vehicle type in comparison to the density estimations calculated from the simulation results using the optimized set of parameters is shown in Fig.~\ref{fig:IV_histogram}.
Generally, the same minimum was found for the majority of the estimation runs for all four parameters, confirming the robustness of our optimization framework, see Fig.~\ref{fig:IV_Parameters}.
Some exceptions could be observed in terms of $\sigma\subs{vel,car}$.
Here, our optimization framework, depending on the starting values, sometimes overestimated $\sigma\subs{vel,car}$, due to the underlying measured data including some cars with a significantly higher mean velocity ($>$160 km/h).
This could potentially be rectified in the future by including a third behavior type, beside the already included \emph{truck} and \emph{car} vehicles.

The identification of the number of minimum necessary agent classes for the optimal simulative mapping of real traffic can be integrated with the optimization algorithm in the future. After an initial clustering of the data to identify possible agent classes, the amount would be iteratively increased for several optimization runs. The quotient of the resulting different optimization results to the optimum can serve as a measure for significance. If the significance for additional agent classes falls below an acceptance threshold to be defined, the optimal number of agent classes is reached.
Yet, as the data we have been using as part of the optimization is only a very small subset of the expected future available data, such an optimization thus far lacks a sufficient data basis and would be prone to overfitting.

\subsection{Extension of Optimization}
Up to now, the main focus of the simulations has solely been on adjusting the four parameters describing the distributions of the \emph{desired velocity} via optimization.
Yet, in order to recreate a more complete picture of the traffic dynamics, our framework needs to be expanded to include the other parameters of the behavior model as well.
Thus, distributions of metrics, which correlate with specific parameters of the Wiedemann99 model, need to be defined.
These distributions can then be added to the definition of the objective function \eqref{eq:objective}, to broaden the scope of the optimization framework.
As part of this project, a further focus will be set on metrics that ensure the inclusion of realistic critical traffic situations within the simulated dynamics.

\begin{figure}
    \centering
    \includegraphics[width=0.49\textwidth]{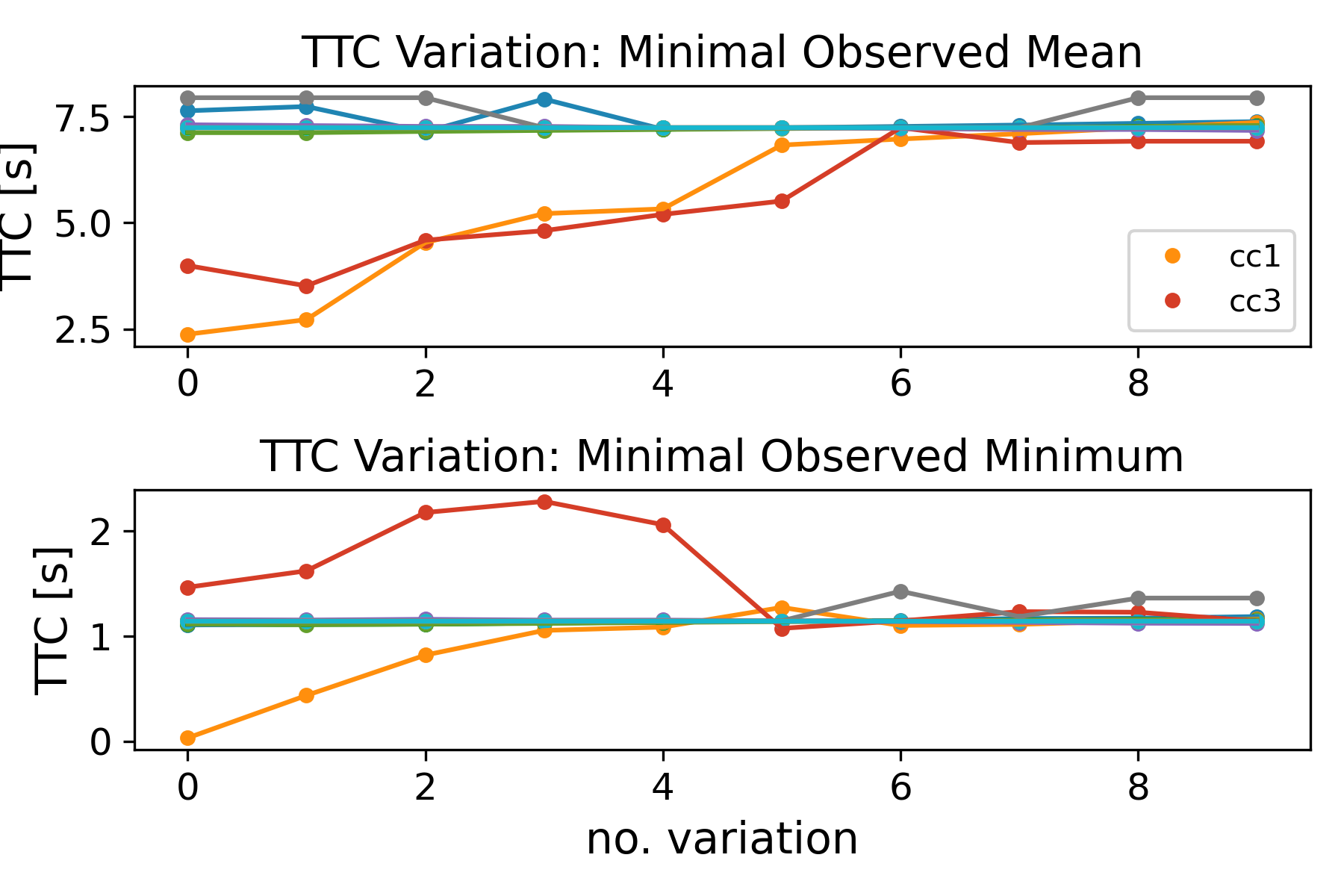}
    \caption{Variation of the minimal observed mean (top) and minimal observed minimum (bottom) \emph{time to collision} (TTC) values when varying Wiedemann99 parameters for a subset of car agents (different colors encode different parameter variations). Only the parameters cc1 (orange) and cc3 (red) show significant influences.}
    \label{fig:WiedemannPara}
\end{figure}

To illustrate this for one such potential metric, Fig.~\ref{fig:WiedemannPara} presents a sensitivity analysis regarding the influence of Wiedemann99 parameters on the aforementioned \emph{time to collision} (TTC) metric, i.e., the ratio of the distance between consecutive cars and their velocity difference \cite{westhofen2023criticality}.
The data inspected here consist of numerous simulations with varying Wiedemann99 parameters \cite{VissimWiedemann2023} applied to a subset of simulated cars. Each of the ten Wiedemann99 parameters (cc0--cc9) was varied individually within the value ranges listed in Tab.~\ref{tab:Wiedemann99para}, while fixing the other parameters to PTV Vissim default values.
Every single  simulation has a duration of 4~hours and the traffic composition is similar to the data considered for the optimization (Fig.~\ref{fig:IV_histogram}), i.e., consisting of roughly $80~\%$ cars and $20~\%$ trucks.

\begin{table}
\centering
\caption{Variation of Wiedemann99 parameters.}
\label{tab:Wiedemann99para}
\begin{tabular}{@{}cccc@{}}
\toprule
Parameter & Start Value & End Value & Default Value \cite{VissimWiedemann2023} \\ \midrule
cc0 & $0.25$ m & $2.5$ m & $1.5$ m \\
cc1 & $0.1$ s & $1.0$ s & $0.9$ s \\
cc2 & $1.0$ m & $5.5$ m & $4.0$ m \\
cc3 & $-2.0$ s & $-11.0$ s & $-8.0$ s \\
cc4 & $-0.1$ m/s & $-0.55$ m/s & $-0.35$ m/s \\
cc5 & $0.1$ m/s & $0.55$ m/s & $0.35$ m/s \\
cc6 & $8.44$ (m$\cdot$s)$^{-1}$ & $12.94$ (m$\cdot$s)$^{-1}$ & $11.44$ (m$\cdot$s)$^{-1}$ \\
cc7 & $0.1$ m/s$^2$ & $0.55$ m/s$^2$ & $0.25$ m/s$^2$ \\
cc8 & $1.0$ m/s$^2$ & $5.5$ m/s$^2$ & $3.5$ m/s$^2$ \\
cc9 & $0.5$ m/s$^2$ & $5.0$ m/s$^2$ & $1.5$ m/s$^2$ \\
\bottomrule
\end{tabular}
\vspace{-10pt}
\end{table}

Fig.~\ref{fig:WiedemannPara} depicts the minimal observed mean TTC and the minimal observed minimum TTC extracted from the simulations for the cars with altered Wiedemann parameters.
Here, the minimal mean TTC refers to the single altered vehicle with the smallest mean TTC per simulation, while the smallest minimal TTC refers to the smallest TTC that was observed in any timestep of the respective simulation.
It can be seen that smaller values of cc1, which represents the preferred gap time of a car to the preceding vehicle, lead to a smaller TTC (Fig.~\ref{fig:WiedemannPara}, orange curves), indicating that the simulated traffic has become more critical.
Additionally we can observe that parameter cc3, which relates to the time where the agent starts to decelerate when approaching a slower vehicle, also has an influence (Fig.~\ref{fig:WiedemannPara}, red curves).
Consequentially, in the future, the potential inclusion of the distribution of the TTC would allow to expand our previously illustrated framework to incorporate the optimization of cc1 and cc3.

\section{USE CASES}
For the productive usage of the developed behavior models and OpenDRIVE scene descriptions, PTV Vissim will be coupled with a Carla-based simulator. Carla is an open-source driving simulator that offers a detailed and realistic virtual environment based on the Unreal Engine \cite{unrealengine} for testing automated driving systems as SUTs. It supports a wide range of sensors and offers customization options, making it a versatile tool for testing and development. 

While PTV Vissim will take over the simulation of the surrounding traffic, the Carla-based simulator will be responsible for simulating the vehicle with the SUT deployed. It features a digital twin of the recording vehicle described in \cite{Haselberger-Jupiter-2022}. The digital twin allows the SUT to be integrated into the simulation loop by incorporating an accurate vehicle dynamics model, sensor model and vehicle interfaces. Based on this, the twin will be used not only for the resimulation of the recorded scenarios, but also for the evaluation of the SUT in different scenarios and scenes based on the behavioral models in PTV Vissim established according to Sec.~\ref{sec:model-parametrization}. Here, we will use a prototypical adaptive cruise control (ACC) advanced driver assistance system and a driver-assisted parking function as exemplary SUTs for the  evaluation of the virtual validation environment.

The full pipeline can then be applied to various use cases in the field of automated driving function development and validation. With this setup, the SUT becomes testable before the first prototype of the vehicle platform is available. Moreover, the number of costly real-world tests, requiring for example exclusive access to a test track (where the degree of realism again is questionable for a different set of reasons), can be reduced, as a pre-calibration of the parameters can be performed in the virtual environment. In addition to that, testing of the SUT is easily scalable and an exact repetition of tests is enabled, e.g., to ensure consistency between different versions. 
Due to the variety of acquisition domains, also a variety of automated driving functions can be tested, ranging from ACC functions operating on highways to parking functions. 
 
A key benefit of using behavioral models in combination with simulation is the capability to test scenarios across a variety of environments. The transferability of these models allows the evaluation of SUTs in different scenes, such as different variations of lane guidance on highways. Furthermore, this approach offers the potential to account for regional disparities in driving behavior. For instance, driving behavior in Europe may vary from that in China  \cite{UZUMCUOGLU202087}, making a European behavior model more precise when simulating traffic scenarios and ultimately assessing SUTs in Europe. This ability to incorporate regional variations in driving behavior can produce more accurate evaluations of SUTs in diverse locations.

\section{CONCLUSIONS AND OUTLOOK}\label{sec:conclusion}

We have presented an approach to systematically acquire real-world data of driving behavior under critical conditions, and utilizing this data to parameterize established driving behavior models, outlining both systematic acquisition goals and assumptions, as well as technical methods for extracting and tracking traffic participants with sufficient accuracy and optimizing the behavior models based on the established Wiedemann99 model framework.

The real-world data are acquired at accident hotspots with homogeneous accident types, using a variety of acquisition means with individual strengths and limitations. Individual processing chains transform the data into a common format that can be processed consistently without the need to consider differences in acquisition methods. Method-dependent data are preserved, however, through optional parameters---for example the state of vehicle lights, especially turn indicators, or interior/passenger data.

The analyses indicate that a defined minimal set of parameters, which can be provided by all considered acquisition methods (including existing datasets based on accident reconstruction), is sufficient for parametrizing microscopic behavior models towards a more realistic behavior under critical conditions, and can serve to provide a more realistic and data-based foundation for the virtual evaluation of automated driving functions.

\subsection*{Outlook}

As part of the AVEAS research project, aiming at the systematic acquisition of behavior data for use in the virtual validation of automated driving functions, the presented results only provide a proof of concept. The amount of data required to satisfy the requirements for safety assurance of SAE Level 4 or 5 functions in complex environments can only be provided through a long-term and international acquisition effort. To achieve this, the raw data processing methods must significantly reduce the effort for manual post-processing and provide means of determining \emph{quantitatively} the statistical significance of the acquired data.

To support international cooperation on this challenge, a recommended format definition based on OpenSCENARIO, OpenDRIVE and OpenLABEL is underway in cooperation with ASAM e.V. that can be used across related project efforts for road safety data acquisition.

The significance of extracted behavior models based on this data still remains to be shown in the aforementioned quantitative sense, including the comparison of the extracted models to larger quantities of data, a systematic identification of distinct behavior models, and an analysis as to whether the ``near miss'' definition can be supported stochastically by approximating the statistical accident count over time. Based on the paradigm of data-driven, stochastic validation, only a quantitative measure of quality for the extracted virtual validation pipeline can enable simulations to be used \emph{in lieu} of real-world test drives.

\section*{ACKNOWLEDGEMENT}

Dedicated to Alexander Muckenhirn (\textborn{}1978, \textdied{}2023), our long-time pilot for the aerial image acquisition, without whom we would not have started this project together.

\bibliographystyle{IEEEtran}
\bibliography{IEEEabrv,root}

\end{document}